\documentclass[conference,letterpaper]{IEEEtran}

\usepackage[T1]{fontenc}
\usepackage{lmodern}

\usepackage{graphicx}
\usepackage{amsmath,amssymb}
\usepackage{booktabs}
\usepackage{algorithm}
\usepackage{algpseudocode}
\usepackage{balance}
\usepackage{url}

\usepackage{url}

\begin{document}

\title{A Graph-Based Approach to Spectrum Demand Prediction Using Hierarchical Attention Networks}

\author{%
\IEEEauthorblockN{%
\mbox{Mohamad Alkadamani\IEEEauthorrefmark{1}\IEEEauthorrefmark{2}}, %
\mbox{Halim Yanikomeroglu\IEEEauthorrefmark{2}}, %
\mbox{Amir Ghasemi\IEEEauthorrefmark{1}}}%
\IEEEauthorblockA{\IEEEauthorrefmark{1}Department of Data Science, Communications Research Centre Canada}%
\IEEEauthorblockA{\IEEEauthorrefmark{2}Department of Systems and Computer Engineering, Carleton University}%
\IEEEauthorblockA{Email: mohamad.alkadamani@ised-isde.gc.ca, amir.ghasemi@ised-isde.gc.ca, halim@sce.carleton.ca}%
}

\maketitle

\begin{abstract}
The surge in wireless connectivity demand, coupled with the finite nature of spectrum resources, compels the development of efficient spectrum management approaches. Spectrum sharing presents a promising avenue, although it demands precise characterization of spectrum demand for informed policy-making. This paper introduces HR-GAT, a hierarchical resolution graph attention network model, designed to predict spectrum demand using geospatial data. HR-GAT adeptly handles complex spatial demand patterns and resolves issues of spatial autocorrelation that usually challenge standard machine learning models, often resulting in poor generalization. Tested across five major Canadian cities, HR-GAT improves predictive accuracy of spectrum demand by 21\% over eight baseline models, underscoring its superior performance and reliability.
\end{abstract}


\section{Introduction}

The increasing reliance on mobile broadband, alongside advancements towards sixth-generation (6G) networks, has significantly intensified competition for radio spectrum resources \cite{matinmikko2020spectrum}. With mobile data traffic projected to nearly triple by 2030, the necessity for efficient spectrum management strategies is more pressing than ever \cite{ericsson2019mobility}. The heterogeneous nature of next-generation networks further necessitates dynamic spectrum access models to meet diverse service requirements \cite{article12}. In this regard, spectrum sharing has emerged as a crucial strategy, enabling more efficient utilization by granting access to underused bands across various stakeholders \cite{baldini2020ml}. However, the efficacy of such frameworks heavily depends on the accurate estimation and localization of spectrum demand, which ensures optimal allocation with minimal interference.

Traditional approaches to spectrum demand estimation often hinge on theoretical network models or rely on simple demographic indicators like population density \cite{telecom2020ai}. Frameworks proposed by institutions such as the International Telecommunication Union (ITU) and the Federal Communications Commission (FCC) apply structured methodologies incorporating market data and usage patterns but typically lack the spatial granularity required for precise spectrum management \cite{itu2020, fcc2020}. These methodologies are often supplemented with empirical models \cite{gsma2025} or statistical techniques \cite{wibisono2015}, which, although useful, often rely on assumptions of uniformly distributed demand.

Furthermore, limited access to proprietary traffic data constrains regulators, impeding a comprehensive understanding of real-world spectrum usage trends. In response, there is growing interest in data-driven approaches utilizing open-source datasets, which facilitate the identification of both spectrum congestion and underutilization zones, thus supporting adaptive and targeted allocation strategies \cite{aygul2025ml}.

Advances in artificial intelligence (AI) present promising alternatives to traditional models, particularly in their ability to reveal latent patterns within large-scale, multi-source datasets \cite{sabir2024ai}. Deep learning approaches, such as convolutional neural networks (CNNs), have been employed for demand estimation, although they often lack mechanisms for explicitly capturing spatial relationships, thereby limiting their effectiveness \cite{10757733}.

Graph neural networks (GNNs) have emerged as particularly effective in this domain, offering robust frameworks for spatial tasks by explicitly modeling relationship and dependencies between neighboring elements \cite{jing, article11}. This capability is paramount in spectrum analysis, where spatial autocorrelation has traditionally degraded the generalization of ML models \cite{app14166989}. Recent studies employing GNNs for spatial regression tasks \cite{hou2024graphconstructionflexiblenodes, Chong} demonstrate significant improvements in performance by effectively capturing geospatial-temporal dependencies.

In light of these advancements, this work present HR-GAT, a Hierarchical-Resolution Graph Attention Network designed to estimate spectrum demand with fine-grained spatial detail. This approach begins with constructing a validated spectrum demand proxy from publicly accessible network deployment records, carefully aligned with measurements from real mobile network operators (MNOs). Multi-resolution geospatial features are then used to build a comprehensive, domain-agnostic feature set for input into HR-GAT's hierarchical graph structure. The key contributions of this work are summarized as follows:
\begin{itemize}
\item \textbf{Development of a validated spectrum demand proxy:} A proxy for mobile spectrum demand is developed by leveraging publicly accessible infrastructure deployment data, validated against actual mobile network operator (MNO) traffic measurements to ensure reliability and accuracy.
\item \textbf{Introduction of hierarchical geospatial modeling via HR-GAT:} An innovative graph attention network is proposed, capturing spatial dependencies across multiple resolutions. 
\item \textbf{Proven generalizability across diverse regions:} HR-GAT is evaluated comprehensively across five major Canadian metropolitan areas, showcasing its strong capacity for generalization and scalability.
\end{itemize}

The remainder of this paper is organized as follows: Section~\ref{sec:problemformulation} formulates the problem; Section~\ref{sec:methodology} outlines the methodology; Section~\ref{sec:proxy validation} presents the proxy validation;  Section~\ref{sec:performance evaluation} presents the model performance evaluation results; and Section~\ref{sec:conclusion} provides concluding remarks.

\section{Problem Formulation}
\label{sec:problemformulation}

This study addresses the estimation of long-term spectrum demand for regulatory purposes, framing it as a supervised learning problem. The primary objective is to predict a spectrum demand proxy using non-technical geospatial features.

The proxy, derived from publicly available deployment data, represents demand levels across geographic areas at the grid tile level. Each city is divided into grid tiles at three resolution levels, creating a hierarchical view of spatial distribution.

For each grid tile \( g \), the problem is mathematically formulated as
\begin{equation}
    y_g = f(\mathbf{x}_g, \mathcal{N}(g), z_g) + \epsilon,
\end{equation}

\noindent where:
\begin{itemize}
    \item \( y_g \): Represents the demand proxy for grid tile \( g \).
    \item \( \mathbf{x}_g \): A feature vector capturing demographic, economic, and social indicators for grid tile \( g \).
    \item \( \mathcal{N}(g) \): The set of neighboring grid tiles, encapsulating spatial dependencies.
    \item \( z_g \): The resolution level of grid tile \( g \), facilitating multi-resolution analysis.
    \item \( \epsilon \): The error term, reflecting unexplained variance in the model's estimation.
\end{itemize}

The goal of the supervised learning task is to uncover the relationship between these indicators and the demand proxy, enabling HR-GAT to predict demand patterns effectively across diverse urban landscapes.

\section{Methodology}
\label{sec:methodology}

This methodology outlines how HR-GAT leverages geospatial data to estimate spectrum demand. By constructing a multi-resolution graph structure, HR-GAT effectively captures spatial dependencies, facilitating accurate predictions across varied urban landscapes. The following subsections detail the graph construction, feature processing, and model architecture.

\subsection{Graph Construction for Multi-Resolution Learning}

Spectrum demand estimation benefits greatly from capturing spatial dependencies via a well-constructed graph \( G = (V, E) \), where:
\begin{itemize}
    \item \( V = \{ v_1, v_2, \dots, v_N \} \), represents the set of nodes each corresponding to a distinct geographic grid tile.
    \item \( E = \{ e_{ij} \} \), represents the connections, or edges, defined between these nodes based on a k-nearest neighbor strategy, capturing spatial locality and dependencies.
\end{itemize}

Each node \( v_g \) embodies a hierarchical embedding defined as
\begin{equation}
    h_g = \sum_{z \in \{13, 14, 15\}} \gamma_z h_g^{(z)},
\end{equation}
where \( h_g^{(z)} \) denotes embeddings at different zoom levels using Bing Maps tiles at zoom levels 13, 14, and 15~\cite{BingMapsTiles}. These levels correspond to different geographic resolutions, capturing intricate spatial details, from broad to fine-grained, offering a rich contextual understanding. The weights \( \gamma_z \) are learnable parameters that determine the influence of information from each resolution.

Edges are assigned weights \( w_{ij} \) using a Gaussian kernel to reflect spatial proximity:
\begin{equation}
    w_{ij} = \exp\left(-\frac{\| s_i - s_j \|^2}{\sigma^2}\right),
\end{equation}
\noindent where \( w_{ij} \) denotes the weight assigned to the edge between nodes \( i \) and \( j \), \( s_i \) and \( s_j \) are the spatial coordinates of these nodes, \(\| s_i - s_j \|^2\) represents their squared Euclidean distance, and \( \sigma \) is a scaling parameter controlling the influence of distance on weight calculation. 

\begin{figure}[t]
\centering
\includegraphics[width=\linewidth]{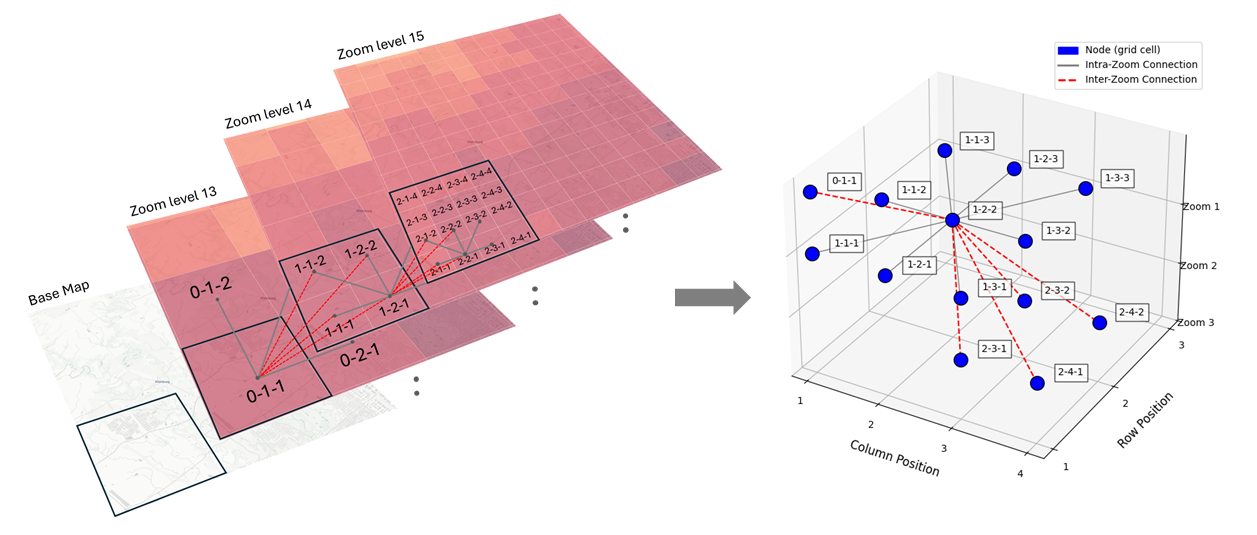}
\caption{Illustration of the HR-GAT hierarchical graph construction.}
\label{fig:hierarchical_graph}
\end{figure}

This ensures that connections between proximate nodes are weighted appropriately, enhancing the model's spatial awareness by emphasizing closer spatial relationships. Fig.~\ref{fig:hierarchical_graph} illustrates the HR-GAT hierarchical graph construction. On the left, the geospatial organization of grid tiles across multiple zoom levels is shown, highlighting how each tile connects based on spatial adjacency and hierarchical relationships. The right side presents the corresponding network representation, where gray solid lines indicate intra-zoom edges, while red dashed lines represent inter-zoom connections across different levels.

\subsection{Feature Processing}
The study utilizes a variety of datasets to create a comprehensive model for spectrum demand estimation. Infrastructure deployment data, along with actual MNO traffic measurements, are utilized to establish a validated spectrum demand proxy, serving as the target variable. Geospatial features are organized onto a grid structure using the Bing Maps tiles at the three levels. These features comprise point-based data, such as specific locations, and polygon-based data, which can include metrics (e.g., census data with statistical attributes) or lack such metrics (e.g., count of building footprints for density measures). By aggregating these features at each grid tile across all zoom levels, consistency is ensured, facilitating accurate comparisons. A detailed summary of the processed features is available in Table~\ref{tab:features_summary}.

 \begin{table}[h]
\centering
\caption{Summary of Features Processed for Demand Modeling}
\label{tab:features_summary}
\renewcommand{\arraystretch}{1.2}
\begin{tabular}{|p{3.5cm}|p{4.3cm}|}
\hline
\textbf{Feature Category} & \textbf{Description} \\ \hline
\textbf{Population Density} & Number of residents per grid tile \\ 
\textbf{Household Density} & Number of households per grid tile \\ 
\textbf{Daytime Population} & Estimated population during working hours \\ 
\textbf{Economic Activity} & Number of businesses by sector \\ 
\textbf{Infrastructure Density} & Number of buildings and their total area \\ 
\textbf{Road Network} & Total road length and number of roads \\ 
\textbf{Transit Accessibility} & Number of transit stops \\ \hline
\end{tabular}
\end{table}

\subsection{HR-GAT Model Architecture}

The HR-GAT model employs attention-based mechanisms to dynamically process and integrate data across multiple spatial resolutions. The architecture is designed to capture and prioritize relevant spatial relationships, enhancing prediction accuracy.

\begin{figure*}[h]
\centering
\includegraphics[width=\linewidth]{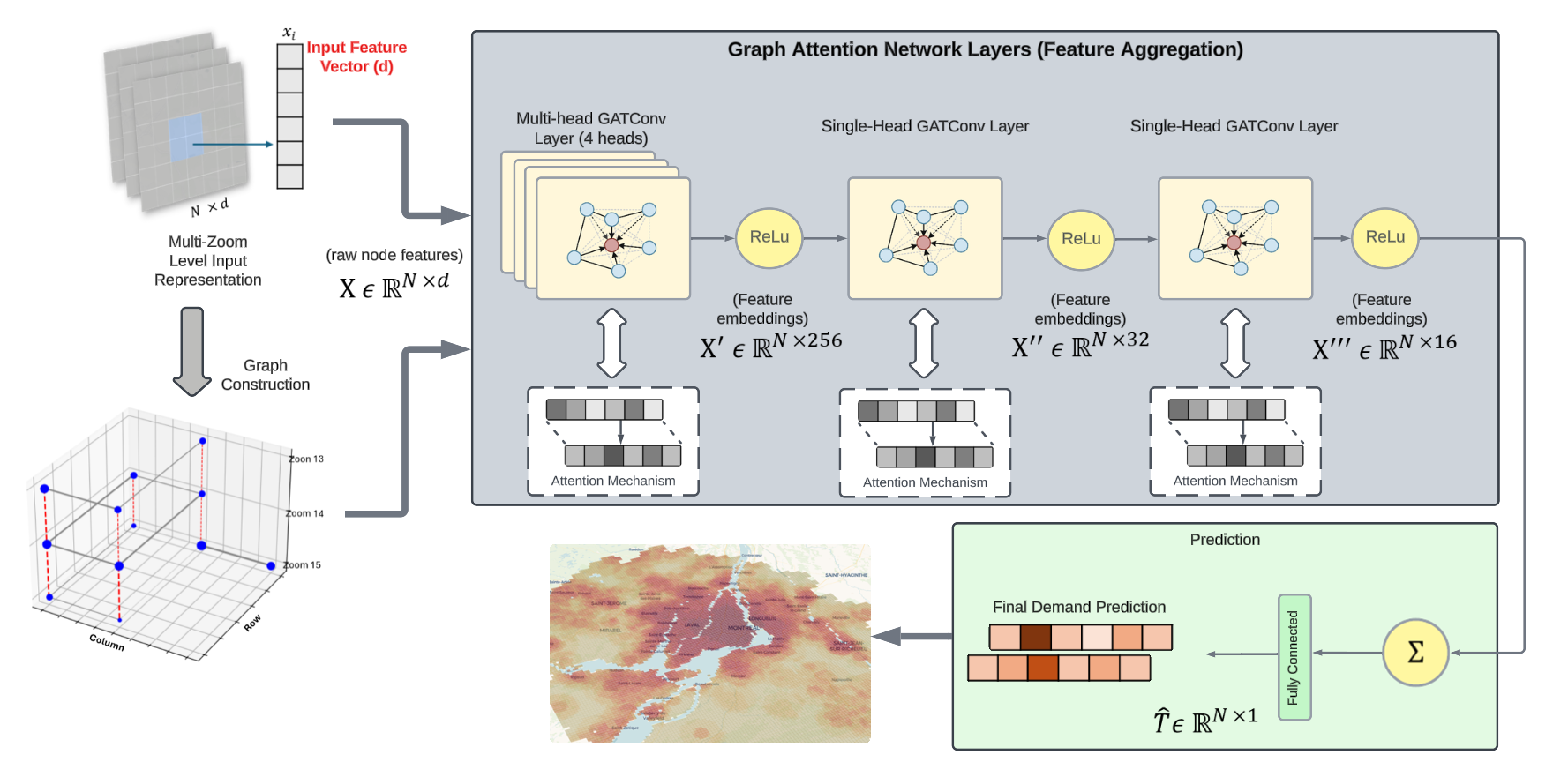} 
\caption{HR-GAT Architecture: Multi-Zoom Spectrum Demand Estimation Model.}
\label{fig:HR-GAT Architecture}
\end{figure*}

The hierarchical attention mechanism in HR-GAT calculates attention scores to modulate the influence of neighboring nodes. These scores are critical for grounding predictions in relevant contextual data:
\begin{equation}
    \alpha_{ij} = \frac{\exp(\text{LeakyReLU}(a^T [Wx_i \Vert Wx_j]))}{\sum_{k \in \mathcal{N}(i)} \exp(\text{LeakyReLU}(a^T [Wx_i \Vert Wx_k]))},
\end{equation}

\noindent where \(\alpha_{ij}\) is the score reflecting node \(j\)'s influence on \(i\), \(\text{LeakyReLU}(\cdot)\) introduces non-linearity, \(a\) is a learnable attention vector, \(W\) is a weight matrix transforming features, and \(x_i, x_j\) are feature vectors for nodes \(i\) and \(j\). \(\Vert\) denotes concatenation, and \(\mathcal{N}(i)\) includes nodes connected to \(i\).

Figure~\ref{fig:HR-GAT Architecture} details the HR-GAT architecture, showing the progression from initial feature inputs through graph convolution operations to aggregated feature embeddings, culminating in precise spectrum demand estimation.

\subsection{Algorithm and Experimental Setup}

The HR-GAT model training involves a series of algorithmic steps designed to refine node embeddings and optimize model performance. The process is detailed in Algorithm~\ref{alg:HR-GAT}. To ensure robust validation and generalization, the model is tested across datasets from five Canadian cities using Clustering-Based Cross-Validation (CBCV) and Leave-One-City-Out (LOCO) validation strategies.

\begin{algorithm}[h]
\caption{HR-GAT Training Algorithm}
\label{alg:HR-GAT}
\begin{algorithmic}[1]
\Require Graph \( G = (V, E) \), node features \( X \), adjacency matrix \( A \), target demand \( T_{\text{grid}} \)
\Require Learning rate \( \eta \), weights \( W^{(l)} \), attention vector \( a \), loss weight \( \lambda \), epochs \( E \)
\State Initialize model parameters
\For{epoch \( e = 1 \) to \( E \)}
    \For{each node \( i \in V \)}
        \State Calculate attention scores to update node embeddings
    \EndFor
    \State Update model parameters using gradient descent
\EndFor
\State \Return: Trained model parameters
\end{algorithmic}
\end{algorithm}

CBCV reduces spatial data leakage by ensuring spatially distinct folds, while LOCO assesses generalization by training on multiple cities and testing on an unseen city. These strategies validate the HR-GAT’s adaptability and robustness in diverse urban environments.

\section{Proxy Validation}
\label{sec:proxy validation}

The evaluation of the spectrum demand proxy begins with validation against actual traffic data obtained from a major mobile network operator (MNO). The LTE traffic dataset spans 15 days and includes hourly downlink throughput for 2,772 LTE cells in Ottawa. This dataset serves as the ground truth, essential for developing a reliable spectrum demand proxy.

For each LTE cell, peak hour download throughput \(T_{\text{DL}, i}\) is recorded and averaged over the 15-day period to represent the maximum load on the cell:
\begin{equation}
    \bar{T}_{\text{DL}, i} = \frac{1}{D} \sum_{d=1}^{D} \max_{h} \left( T_{\text{DL}, i, h, d} \right),
\end{equation}

\noindent where \( \bar{T}_{\text{DL}, i} \) is the average peak hour throughput for cell \(i\), \( D \) is the number of days in the observation period (15 days), \( T_{\text{DL}, i, h, d} \) is the downlink throughput of cell \(i\) at hour \(h\) on day \(d\), and \( \max_{h}(\cdot) \) identifies the maximum hourly throughput for each day.

To establish a spatial representation, a propagation model is applied using power levels and antenna heights obtained from the data. The enhanced Hata model calculates coverage, assuming typical mobile receive sensitivity. Coverage for each cell is mapped to grid tiles, and traffic is allocated as follows:
\begin{equation}
    T_{\text{DL, grid}}(g) = \sum_{i \in \mathcal{C}(g)} \frac{T_{\text{DL}, i}}{|\mathcal{G}(i)|},
\end{equation}
\noindent where \( T_{\text{DL, grid}}(g) \) is the total downlink traffic assigned to grid tile \(g\), \( \mathcal{C}(g) \) is the set of cells that cover tile \(g\), \( T_{\text{DL}, i} \) is the peak hour throughput for cell \(i\), and \( |\mathcal{G}(i)| \) is the number of grid tiles covered by cell \(i\), used to evenly distribute the cell's traffic.

The deployed bandwidth \(\mathsf{BW}_{\text{grid}}\) is similarly allocated to ensure an accurate spatial proxy:
\begin{equation}
    \mathsf{BW}_{\text{grid}} = \sum_{i \in \mathcal{C}(g)} \frac{\mathsf{BW}_i}{|\mathcal{G}(i)|}.
\end{equation}

\noindent where \( \mathsf{BW}_i \) is the bandwidth deployed at cell \(i\), and the remaining parameters are as previously defined.

\begin{figure}[t]
\centering
\includegraphics[width=\linewidth]{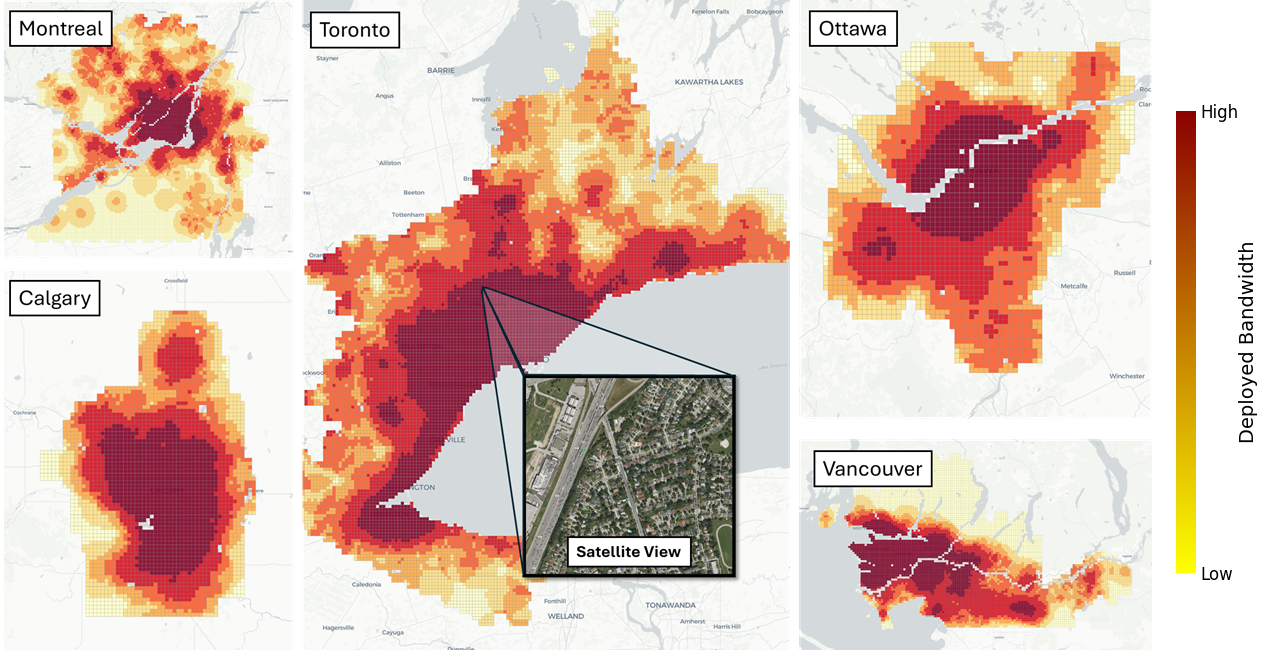}
\caption{Deployed bandwidth heatmaps at zoom level 15 for Calgary, Montreal, Toronto, Vancouver, and Ottawa. Darker regions indicate higher estimated spectrum deployment.}\label{fig:heatmap}
\end{figure}

Figure~\ref{fig:heatmap} displays bandwidth heatmaps for the targeted cities, visually confirming regional demand variation.

Validation of the proxy is achieved through Ordinary Least Squares (OLS) regression analysis, comparing the proxy with the LTE traffic data. The analysis yields an R-squared value of 0.727, confirming the proxy's reliability, as presented in Table~\ref{tab:ols_results}. This validation ensures that subsequent steps are based on a thoroughly vetted proxy, accurately representing spectrum demand.

\begin{table}[t]
\centering
\caption{OLS Regression Results for Proxy Validation}
\label{tab:ols_results}
\begin{tabular}{lccc} 
\toprule
\textbf{Proxy Variable} & \textbf{R²} & \textbf{F-Statistic} & \textbf{p-Value} \\ 
\midrule
Deployed Bandwidth (PBW) & 0.727 & 4477 & $\mathbf{< 0.001}$ \\ 
\bottomrule
\end{tabular}
\end{table}

\section{ Performance Evaluation}
\label{sec:performance evaluation}

The performance evaluation of HR-GAT was conducted using data encompassing 50,679 grid tiles across three zoom levels (13, 14, and 15) for the five selected Canadian cities. The model utilized a final set of 30 geospatial features selected for their impact. Two rigorous validation strategies were employed: Clustering-Based Cross-Validation (CBCV) and Leave-One-City-Out (LOCO).

\subsection{Cross-Validation Results (CBCV)}
Using a five-fold CBCV approach within each city, HR-GAT was compared against eight baseline models. The results, summarized in Table~\ref{tab:model_comparison}, demonstrate HR-GAT's superior performance. It achieved the lowest median MAE (10.93), lowest median RMSE (29.30), and highest R² (0.91). This highlights the effectiveness of modeling spatial adjacency and hierarchical structure. Plain GAT, while utilizing graph connectivity, underperformed compared to HR-GAT, likely due to its lack of multi-resolution capabilities. Boosting models (XGBoost, Gradient Boosting) performed well but lacked explicit spatial awareness, while traditional ML models (Decision Tree, Random Forest, Linear Model) and Vanilla CNN showed significantly lower accuracy.

\begin{table}[h]
\centering
\caption{Performance Comparison using CBCV}
\label{tab:model_comparison}
\begin{tabular}{|l|c|c|c|}
\hline
\textbf{Model} & \textbf{Median MAE} ↓ & \textbf{Median RMSE} ↓ & \textbf{R²} ↑ \\ 
\hline
Gradient Boosting & 23.44 & 44.27 & 0.86 \\ 
XGBoost & 23.53 & 44.72 & 0.86 \\ 
LightGBM & 30.78 & 55.70 & 0.80 \\ 
Decision Tree & 41.60 & 79.92 & 0.53 \\ 
Random Forest & 45.89 & 77.48 & 0.55 \\ 
Linear Model & 55.50 & 89.31 & 0.43 \\ 
\textbf{Vanilla CNN} & \textbf{44.23} & \textbf{69.30} & \textbf{0.63} \\ 
\textbf{Plain GAT} & \textbf{24.88} & \textbf{37.30} & \textbf{0.83} \\ 
\textbf{HR-GAT} & \textbf{10.93} & \textbf{29.30} & \textbf{0.91} \\ 
\hline
\end{tabular}
\end{table}

\begin{figure}[t]
\centering
\includegraphics[width=\linewidth]{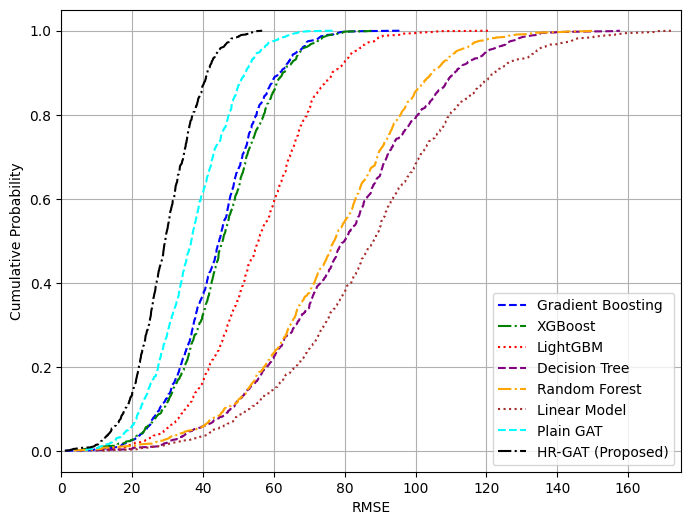}
\caption{eCDF of RMSE values across models. HR-GAT demonstrates superior prediction accuracy.}
\label{fig:ecdf_rmse}
\end{figure}

Figure~\ref{fig:ecdf_rmse} displays the empirical cumulative distribution function (eCDF) of RMSE values, showing HR-GAT's steeper curve which indicates a higher proportion of low-error estimations. 

\begin{figure}[t]
\centering
\includegraphics[width=\linewidth]{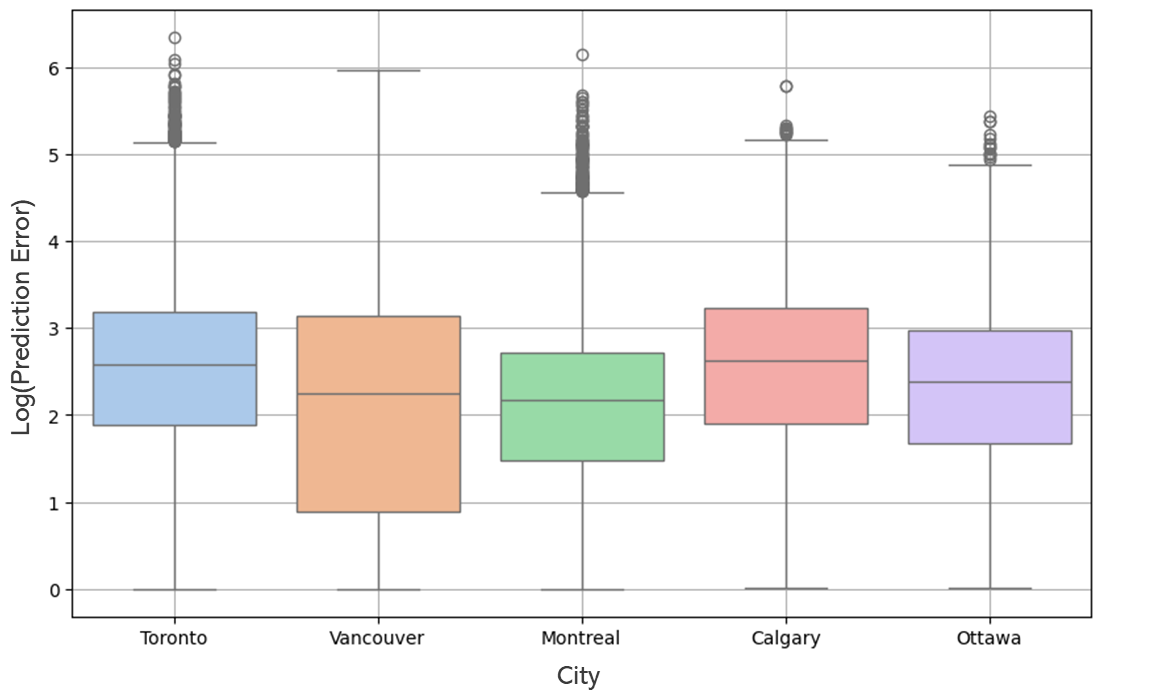}
\caption{Log-scaled residual error distribution across cities, showing HR-GAT's consistent performance.}
\label{fig:residual_boxplot}
\end{figure}

Furthermore, the log-scaled residual error distribution in Fig.~\ref{fig:residual_boxplot} confirms HR-GAT's consistent performance and generalization capability across the different cities studied. The log-scaled residuals indicate that HR-GAT maintains a consistent error distribution across all five cities, suggesting that the model does not exhibit location-specific biases. While minor variations exist due to local urban characteristics, the comparable interquartile ranges and median errors across cities reinforce the model's ability to generalize effectively beyond the training data.

\subsection{Generalization to Unseen City (LOCO)}
The LOCO strategy involved training the model on four cities and testing it on Ottawa, a completely unseen city. As shown in Table~\ref{tab:ottawa_mae}, HR-GAT achieved the lowest median MAE (18.74), significantly outperforming all baselines. This result strongly validates HR-GAT's ability to generalize to new urban environments by effectively leveraging learned hierarchical spatial relationships.

\begin{table}[h]
\centering
\caption{Median MAE for Ottawa (Unseen Test Set)}
\label{tab:ottawa_mae}
\begin{tabular}{|l|c|}
\hline
\textbf{Model} & \textbf{Median MAE (Ottawa)} ↓ \\ 
\hline
Gradient Boosting & 37.34 \\ 
XGBoost & 39.49 \\ 
LightGBM & 47.20 \\ 
Decision Tree & 52.47 \\ 
Random Forest & 50.66 \\ 
Linear Model & 60.93 \\ 
\textbf{Vanilla CNN} & \textbf{48.22} \\ 
\textbf{Plain GAT} & \textbf{23.30} \\ 
\textbf{HR-GAT} & \textbf{18.74} \\ 
\hline
\end{tabular}
\end{table}

\subsection{Key Factors Influencing Spectrum Demand}
Interpreting estimation from complex ML models, such as HR-GAT, requires techniques that explain the contribution of individual features. Since HR-GAT is a black-box model, the SHapley Additive exPlanations (SHAP) method is used to quantify the impact of each feature on the model’s estimations \cite{Lundberg2017}. SHAP assigns an importance value to each feature per estimation, enabling an understanding of how different geospatial, demographic, and economic factors contribute to spectrum demand estimation.

\begin{figure}[h]
    \centering
    \includegraphics[width=0.95\columnwidth]{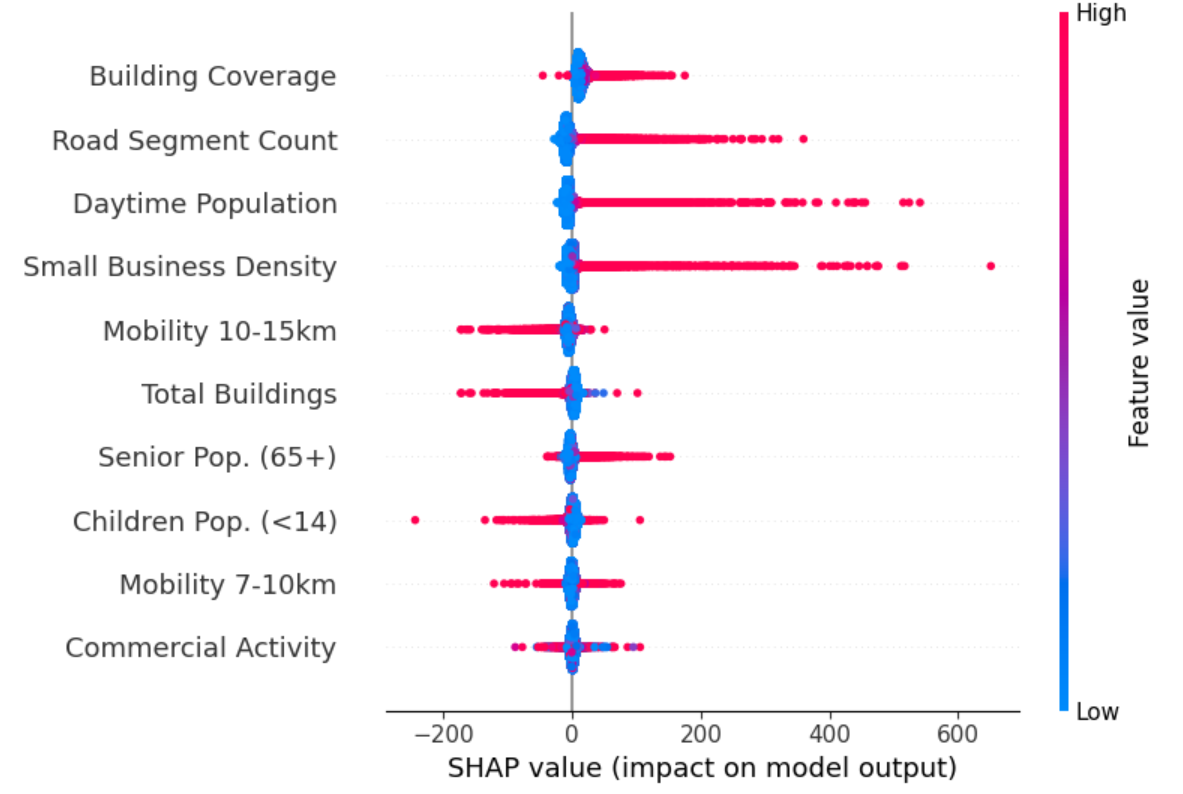}
    \caption{SHAP summary plot illustrating the relative importance of the top 10 features in predicting mobile spectrum demand. Higher absolute SHAP values indicate stronger influence on the model's estimation.}
    \label{fig:shap_summary}
\end{figure}

Figure~\ref{fig:shap_summary} presents the top 10 most influential features out of a total of 30 input features used in the model.

Urban infrastructure and density play a crucial role in shaping spectrum demand. Features such as building coverage, road segment count, and total number of buildings indicate the extent of urban development, with higher values generally correlating with increased mobile traffic due to dense commercial and residential activities. Similarly, daytime population and small business density highlight areas of concentrated human presence, reinforcing the idea that business districts and commercial centers sustain high network loads during peak hours.

Mobility patterns also exhibit a strong influence on spectrum demand. The number of people traveling 7-10 km and 10-15 km captures inter-zonal commuting behaviors, where high movement levels correspond to dynamic variations in network traffic, particularly in transit corridors and suburban-urban interfaces.

Demographic factors provide additional insights into demand variations. The presence of senior citizens (65+) and children ($<$14) contributes to household-driven network usage patterns, albeit with a lower overall influence compared to commercial and commuting-related factors.

Finally, commercial activity, approximated through nighttime luminosity (NTL) \cite{Li2020}, emerges as a significant determinant, capturing economic hotspots where high levels of business and commercial interactions drive sustained mobile service consumption.

These findings demonstrate that spectrum demand is influenced by a combination of urban infrastructure, population distribution, mobility patterns, and economic activity, emphasizing the need for data-driven spectrum allocation strategies.

\section{Conclusion}
\label{sec:conclusion}

This study introduced HR-GAT, a novel hierarchical graph attention network designed for accurate spectrum demand estimation using geospatial data. By explicitly modeling multi-resolution spatial dependencies, HR-GAT effectively captures complex demand patterns and mitigates spatial autocorrelation bias.

Rigorous evaluation across five Canadian cities demonstrated HR-GAT's superior performance, achieving a 21\% predictive accuracy improvement over eight baseline models, with strong generalization capabilities even to unseen urban environments. The model's ability to provide detailed, spatially-aware insights from non-technical indicators offers significant advantages for spectrum management.

Specifically, HR-GAT supports proactive spectrum allocation by identifying high-demand areas, enables advanced dynamic sharing frameworks like DSA, facilitates differentiated strategies for diverse urban/suburban needs, and aids in promoting equitable access by highlighting underserved regions. Ultimately, HR-GAT serves as a valuable data-driven tool, enhancing the efficiency, flexibility, and evidence-based nature of spectrum policy and planning.

\bibliographystyle{IEEEtran} 
\bibliography{references}

\end{document}